\documentclass[10pt,twocolumn,letterpaper]{article}

\usepackage[pagenumbers]{cvpr} 

\usepackage{graphicx}
\usepackage{amsmath}
\usepackage{amssymb}
\usepackage{booktabs}
\usepackage{multirow} 

\usepackage[pagebackref,breaklinks,colorlinks]{hyperref}

\usepackage[capitalize]{cleveref}
\crefname{section}{Sec.}{Secs.}
\Crefname{section}{Section}{Sections}
\Crefname{table}{Table}{Tables}
\crefname{table}{Tab.}{Tabs.}

\def\confName{CVPR}
\def\confYear{2022}

\begin{document}

\title{Coupled Iterative Refinement for 6D Multi-Object Pose Estimation}

\author{Lahav Lipson \quad Zachary Teed \quad Ankit Goyal \quad Jia Deng\\
 Princeton University
}
\maketitle
\begin{abstract}
    We address the task of 6D multi-object pose: given a set of known 3D objects and an RGB or RGB-D input image, we detect and estimate the 6D pose of each object. We propose a new approach to 6D object pose estimation which consists of an end-to-end differentiable architecture that makes use of geometric knowledge. Our approach iteratively refines both pose and correspondence in a tightly coupled manner, allowing us to dynamically remove outliers to improve accuracy. We use a novel differentiable layer to perform pose refinement by solving an optimization problem we refer to as Bidirectional Depth-Augmented Perspective-N-Point (BD-PnP). Our method achieves state-of-the-art accuracy on standard 6D Object Pose benchmarks. Code is available at \url{https://github.com/princeton-vl/Coupled-Iterative-Refinement}.
\end{abstract}

\section{Introduction}

Given an RGB or RGB-D image containing a set of object instances of known 3D shapes, 6D multi-object pose is the task of detecting and estimating the 6D pose---position and orientation---of each object instance. Accurate poses are important for robotics tasks such as grasping and augmented reality applications involving shape manipulation.

In the standard 6D multi-object pose setup, we are given a set of 3D models of known object instances. Given an RGB or RGB-D input image, the goal is to jointly detect object instances and estimate their 6D object pose. Early work solved this problem by first estimating correspondences between the 3D model and the image~\cite{lowe1999object}, producing a set of 2D-3D correspondences, which are then used to obtain 6D object pose using Perspective-n-Point (PnP) solvers~\cite{horaud1989analytic,epnp} or iterative algorithms like Levenberg-Marquardt. 

While 2D-3D correspondence is sufficient to solve for 6D pose, it is difficult to obtain accurate correspondence in practice. In many applications, we wish to estimate the pose of poorly textured objects where local feature matching is unreliable. Furthermore, problems such as heavy occlusion, object symmetry, and lighting variation can make detecting and matching local features near impossible. These problems cause classical systems to be too brittle for many use cases which require a greater degree of robustness.

Recently, many of these issues have been partially addressed using deep learning. A simple approach is to train a network to directly regress 6D poses\cite{cosypose,li2018deepim,xiang2017posecnn}. Direct pose regression simply learns to map input to output, and makes no use of the fact that the pixels are a perspective projection of a known 3D object. Although direct pose regression can be quite effective in practice, an intriguing question is whether there exist better deep learning methods that take advantage of projective geometry.

Many works on 6D pose have attempted to combine deep learning and projective geometry. One approach is to train a deep network to detect keypoints of a known 3D object~\cite{pix2pose,tekin2018real,bb8,pavlakos20176,pvn3d, peng2019pvnet}, producing a set of 2D-3D correspondences which can serve as input to a Perspective-n-Point (PnP) solver. Another approach is to impose geometric knowledge in the form of implicit or declarative layers~\cite{blindpnp,bpnp}. These works showed that PnP could be implemented as a modular component in end-to-end differentiable architectures. However, both approaches are ``one-shot'' in the sense that correspondence is predicted once and then used to solve for pose through a PnP solver (differentiable or not);  this makes the approaches sensitive to outliers and errors in the correspondences.

\begin{figure}[t]
    \centering
	\includegraphics[width=.9\linewidth]{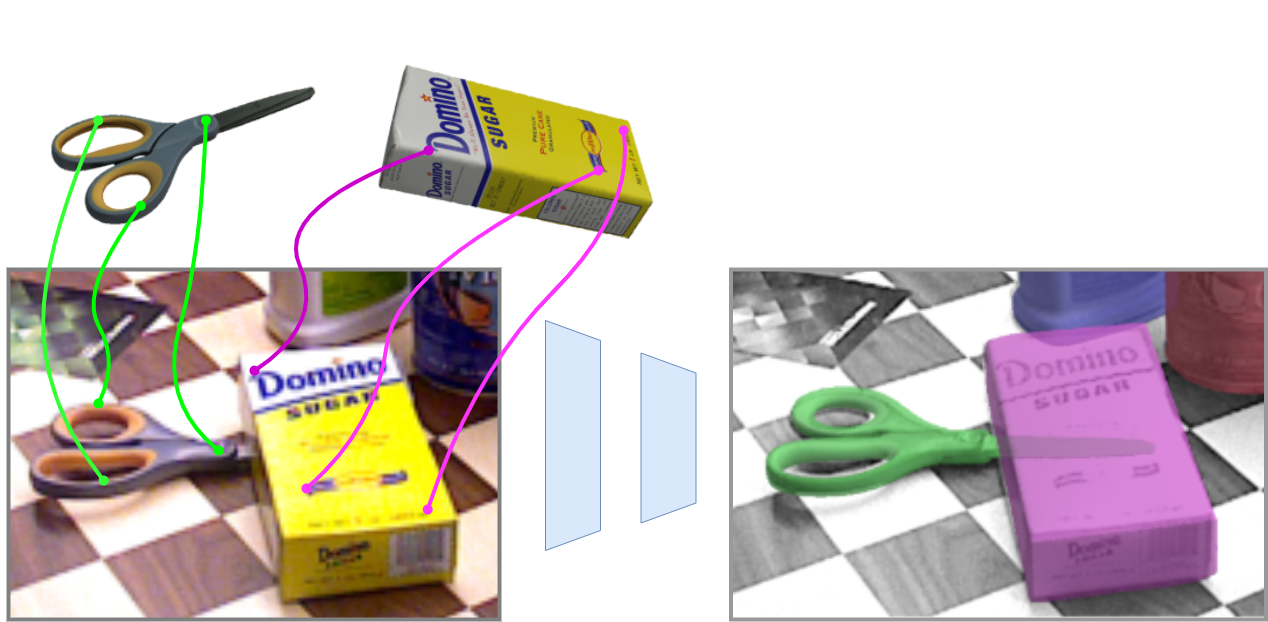}
	\caption{Given an image and collection of 3D models, our method outputs the position and orientation of each object instance.}
	\label{fig:per_obj}
\end{figure}

We propose a new approach to 6D object pose estimation. Our approach consists of an end-to-end differentiable architecture that makes use of geometric knowledge. The main novelty of our approach over prior work on 6D pose is the use of ``coupled iterative refinement'': unlike prior work which operates in a single shot setting, we iteratively refine pose and correspondence in a tightly coupled manner, allowing us to dynamically remove outliers to improve accuracy. 

Our approach builds on top of the RAFT \cite{teed2020raft} architecture developed for optical flow (i.e.\@ dense correspondence). The basic idea is to estimate flow between the input image and a set of rendered images of the known 3D object, generating 2D-3D correspondences that are used to solve for pose. Like RAFT, we use a GRU to perform recurrent iterative updates, but at each iteration we update not only flow but also object pose. The flow update and pose update are tightly coupled: the flow update is conditioned on the current pose, and the pose update is conditioned on the flow. 

To perform the pose update, we introduce a novel differentiable layer we call ``Bidirectional Depth-Augmented PnP (BD-PnP)''. This layer is similar to a differentiable PnP solver in that it produces a Gauss-Newton update to object pose by minimizing reprojection error. However, it is novel in two aspects.  First, it is bidirectional: it solves for a single pose update to simultaneously satisfy two sets of 2D-3D correspondences, one set defined on the input image, the other set defined on a rendered image. Second, our layer is ``depth-augmented'': the optimization objective also includes the reprojection error on inverse depth, which we show to be important for improving accuracy.

Our method achieves state-of-the-art accuracy on the YCB-V \cite{ycb}, T-LESS\cite{tless} and Linemod (Occluded) \cite{linemod_occlusion} RGB-D multi-object BOP \cite{bopchallenge} pose benchmarks, significantly outperforming prior work. A variant of our method can handle RGB-only input, with performance on par with the current state-of-the-art.

\section{Related Work}

\noindent \textbf{Classical Approaches} Early works on 6D object pose estimation used invariant local features~\cite{sift,surf} to generate correspondences between 2D image features and 3D model features~\cite{lowe2001local}. Given the set of 2D-3D correspondences, PnP solvers are then used to estimate 6D object pose, that is, the position and orientation of the object in world coordinates~\cite{gao2003complete}. Both closed form~\cite{gao2003complete,epnp,zheng2013revisiting} and iterative algorithms~\cite{lm} exist for recovering pose from correspondence. In practice, it is common to use a closed form solution as initialization followed by iterative refinement\cite{epnp}. Due to the presence of outliers, robust estimation techniques such as RANSAC~\cite{ransac} are typically required. Local features perform well on highly textured objects, but often fail to produce a sufficient number of accurate correspondences on textureless objects.  

In this work, we also estimate correspondence between the 3D model and the input image to produce a set of 2D-3D correspondences. However, instead of predicting a set of sparse matches, we predict dense correspondence fields between the input image and rendered views of the 3D model together with per-pixel confidence weights. By predicting dense correspondence, we can ensure a sufficient number of matches allowing us to solve for accurate pose even on textureless objects where classical methods fail. 

\smallskip\noindent\textbf{Learning-based Approaches} Several works propose to estimate pose by directly regressing rotation and translation parameters~\cite{posecnn,posenet,do2019real}. 

Other works generate 2D-3D correspondences by detecting or regressing keypoints. One type of keypoint parameterization is object coordinates~\cite{bpnp,pix2pose,brachmann2014learning,bb8}. Given a canonical pose of an object, the object coordinates represent the position of a 3D point in the coordinate system of the canonical pose. Brachmann et al.~\cite{brachmann2014learning} showed that a random forest could be used to regress object coordinates from image features. Pix2Pose~\cite{pix2pose} uses a neural network to regress object coordinates from the image, while BB8~\cite{bb8} estimates bounding box corners. By regressing object coordinates, these systems produce a dense set of 2D-3D correspondences which can be used to estimate object pose using PnP solvers. BPNP~\cite{bpnp} takes this idea a step further and implements the PnP solver as a differentiable network layer. During training, BPNP uses the implicit function theorem to backpropagate gradients through the PnP solver such that the full system can be trained end-to-end. Our work is similar to these approaches in the sense that we also regress 2D-3D correspondences (in the form of optical flow between the input image and rendered views of the 3D model), but we differ by performing coupled iterations where both correspondences and object pose are iteratively refined.

\smallskip\noindent\textbf{Iterative Refinement} It can be challenging to estimate accurate pose in a single-shot setting. This has motivated several works to apply iterative refinement techniques to produce more accurate pose estimates. DeepIM~\cite{li2018deepim} is an iterative ``render-and-compare'' approach to pose estimation. During each iteration, DeepIM uses the current estimate of object pose to render the 3D model, then uses the render and the image to regress a pose update to better align the image with the render. CosyPose~\cite{cosypose} builds on this idea using improved network architectures and rotation paramterizations. 

Similar to DeepIM, our approach also includes an outer loop that re-renders the 3D model using the current pose estimate.  However, our pose updates are produced not by regression but by our BD-PnP layer that makes use of geometric constraints. In particular, the BD-PnP layer solves for a pose update based on the current estimate of flow. 

RAFT-3D~\cite{raft3d} applies iterative refinement in the context of scene flow estimation. Like our work, they iterate between optical flow refinement and fitting rigid body transformations. However, RAFT-3D predicts pixelwise transformation fields between pairs of frames using the Dense-SE3 layer, while our work predicts transformations on the object level using our novel BD-PnP layer, which is substantially different from the Dense-SE3 layer.

\section{Approach}

Our method operates on a single input image and produces a set of object pose estimates (Fig. \ref{fig:per_obj}). For simplicity of exposition, we assume RGB-D input unless otherwise noted. Our method can be decomposed into 3 stages: (1) object detection, (2) pose initialization, and (3) pose refinement. The first two stages (object detection and pose initialization) follow the method proposed by CosyPose~\cite{cosypose}. Our primary contribution concerns the pose refinement stage, where we seek to transform the initial coarse pose estimates into refined poses with subpixel reprojection error.

\smallskip\noindent\textbf{Preliminaries}
Given a textured 3D mesh of an object, we can render images and depth maps of the object from different viewpoints using PyTorch3D~\cite{pytorch3d}, with views parameterized by intrinsic and extrinsic parameters
\begin{equation}
    \mathbf{G}_i = \begin{pmatrix} \mathbf{R} & \mathbf{t} \\ \mathbf{0} & 1 \end{pmatrix} \qquad
    \mathbf{K}_i = \begin{pmatrix} 
        f_x & 0 & c_x \\
        0 & f_y & c_y \\
        0 & 0 & 1 \\
    \end{pmatrix}.
\end{equation}
where $\mathbf{G}_{i}$ is the object pose in camera coordinates. Letting $\mathbf{G}_0$ be the pose for the image and $\{\mathbf{G}_1, ..., \mathbf{G}_{N}\}$ be the poses for a set of renders, we can define a function which maps points in a render to points in the image
\begin{equation}
    \mathbf{x}_{i \rightarrow 0}' = \Pi\left(\mathbf{G}_0 \mathbf{G}_i^{-1} \Pi^{-1}(\mathbf{x}_i) \right)
    \label{eqn:forw}
\end{equation}
or from the image to a render
\begin{equation}
    \mathbf{x}_{0 \rightarrow i}' = \Pi\left(\mathbf{G}_i \mathbf{G}_0^{-1} \Pi^{-1}(\mathbf{x}_0) \right)
    \label{eqn:back}
\end{equation}
We use depth-augmented pinhole projection functions $\Pi$ and $\Pi^{-1}$ which convert not just image coordinates of a point but also its \emph{inverse depth} between frames
\begin{equation}
    \Pi(\mathbf{X}) = \begin{bmatrix} 
        X/Z \\
        Y/Z \\
        1/Z
    \end{bmatrix} \ \
    \Pi^{-1}(\mathbf{x}) = \begin{bmatrix} 
        x/d \\
        y/d \\
        1/d
        \end{bmatrix}
    \  \
    \mathbf{x} = \begin{bmatrix} x \\ y \\ d \end{bmatrix}
\end{equation}
where it is assumed pixels coordinates are normalized using the camera intrinsics.

The goal is to solve for pose $\mathbf{G}_0$ such that Eqn~\ref{eqn:forw} correctly maps points between the image and renders. We can return the object pose in world coordinates by inverting $\mathbf{G}_0$.

\smallskip\noindent\textbf{Object Candidate Detection} Given an input image, we first apply Mask-RCNN~\cite{maskrcnn} to generate a set of object detections and associated labels. We use the pretrained Mask-RCNN weights from CosyPose~\cite{cosypose} which were trained on the BOP~\cite{bopchallenge} object classes. We then use the detected bounding boxes to generate crops from the image, segmentation mask, and depth map (in the RGB-D setting). We resize crops to $320 \times 240$ and adjust intrinsics accordingly.

\smallskip\noindent\textbf{Pose Initialization} Following detection, our system operates in parallel for each object candidate. Given an object, we start by generating an initial pose estimate $\mathbf{G}^{(0)}$.

We first compute a translation vector $\mathbf{t}_\text{bbox}$ which aligns the bounding box of the 3D model to the detected object mask, such that the diameter of the mesh aligns with the projected bounding box. We then render the 3D model using the estimated translation and concatenate the render with the image crop.  This input is fed directly to an Resnet-based architecture which regresses a rotation and translation update ($\mathbf{R}$, $\Delta \mathbf{t}$) where rotation is predicted using the continuous 6D parameterization proposed by Zhou et al.~\cite{zhou2019continuity}. The initial pose estimate can be written as a $4\times4$ matrix 
\begin{equation}
    \mathbf{G}_0^{(0)} = \begin{pmatrix} \mathbf{R} & \mathbf{t}_\text{bbox} + \Delta \mathbf{t} \\ \mathbf{0} & 1 \end{pmatrix}.
\end{equation}

We use the pretrained EfficientNet~\cite{tan2019efficientnet} weights from Cosypose~\cite{cosypose} for this pose initialization step.

\smallskip\noindent\textbf{Feature Extraction and Correlation} Given our initial pose estimate, we render several viewpoints at our pose estimate as well as centered around it by adding or subtracting $22.5^{\circ}$ from either pitch, yaw or roll (7 rendered views in total). For each render, our network estimates bidirectional, dense correspondence between the render and the image crop.  The object pose of each of the renders is known; the pose of the object in the image crop needs to be estimated.

For all $N$ renders we extract dense $\frac{H}{4} \times \frac{W}{4}$ feature maps. We also apply the same feature extraction network to the image crop using shared weights.

We then build two correlation volumes for each image-render pair, one from the image to the render and another from the render to the image. The correlation volume is computed by taking the dot product between all pair of feature vectors. Like RAFT~\cite{teed2020raft}, we pool the last two dimension of each correlation volume to produce a set of 4-level correlation pyramids. These pyramids contain correlation features useful for matching.

\subsection{Coupled Iterative Refinement}

\begin{figure}[t]
    \centering
	\includegraphics[width=.9\linewidth]{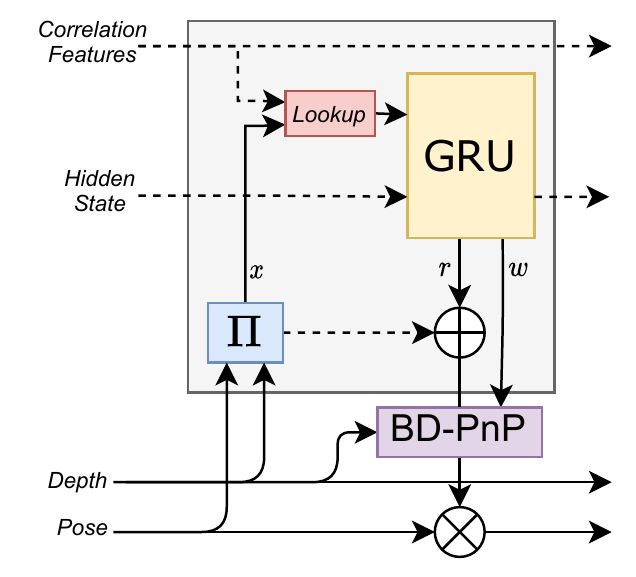}
	\caption{The update operator. A GRU produces revisions $\mathbf{r}$ and confidence weights $\mathbf{w}$. The revisions and confidence weights are used to solve for a pose update.}
	\label{fig:update_op}
\end{figure}

We use a GRU-based update operator (Fig.~\ref{fig:update_op}) to produce a sequence of updates to our pose estimates.  The GRU also has a hidden state which gets updated with each iteration.

Let $\mathbf{G}$ be the set of all poses, including both the renders and the image. The poses of the renders are fixed, while the first pose $\mathbf{G}_0$, the pose of the image, is a variable.

Using Eqn.~\ref{eqn:forw}, we compute the dense correspondence field bidirectionally between the image and each render. We compute $\mathbf{x}_{i\rightarrow 0}$ using Eqn.~\ref{eqn:forw} and $\mathbf{x}_{0\rightarrow i}$ using Eqn.~\ref{eqn:back}. The correspondence field $\mathbf{x}_{i\rightarrow0} \in \mathbb{R}^{H \times W \times 3}$ tells us, for every pixel in render $i$, its estimated 2D location in the image. It is worth noting that the correspondence field is augmented with inverse depth,  that is, $\mathbf{x}_{i\rightarrow0}$ contains not just 2D coordinates but also inverse depth.

\smallskip\noindent\textbf{Correlation Lookup} We use $\mathbf{x}_{i\rightarrow 0}$ to index from the corresponding correlation pyramid using the lookup operator defined in RAFT~\cite{teed2020raft}. The lookup operator constructs a local grid around each point with radius $r$ and uses the grid to index from each level in the correlation pyramid, producing a total of $L$ correlation features. The result of the lookup operation is a map of correlation features $\mathbf{s}_{i\rightarrow 0} \in \mathbb{R}^{H\times W \times L}$. Similarly, we use $\mathbf{x}_{0\rightarrow i}$ to produce the correlation features $\mathbf{s}_{0\rightarrow i}\in \mathbb{R}^{H\times W \times L}$.

\smallskip\noindent\textbf{GRU Update} For each image-render pair, the correlation features $\mathbf{s}_{i\rightarrow 0}$ and the hidden state $\mathbf{h}_{i\rightarrow 0}$, together with additional context and depth features described in the appendix, are fed to a 3x3 convolution GRU, which outputs (1) a new hidden state, 
(2) revisions $\mathbf{r}_{i\rightarrow 0} \in \mathbb{R}^{H \times W \times 3}$ to each of the dense correspondence fields, and (3) a dense map of confidence $\mathbf{w}_{i\rightarrow 0}$ in the predicted revisions. The revision $\mathbf{r}_{i\rightarrow 0}$ represents a new flow estimate in the form of a dense map of corrections that should be applied to the correspondences produced by the current pose estimate. Note that $\mathbf{r}_{i\rightarrow 0}$ includes not just revisions for 2D coordinates but also revisions for inverse depth. The revisions for depth are necessary to compensate for the fact that the input sensor depth may be noisy and the corresponding point may be occluded. 

We also apply the same GRU for the other direction of the image-render pair. That is, we use the correlation features $\mathbf{s}_{0\rightarrow i}$ to produce revisions $\mathbf{r}_{0\rightarrow i}$ and confidence map $\mathbf{w}_{0\rightarrow i}$. Note that the weights of the GRU are shared across all image-render pairs in both directions.

\smallskip\noindent\textbf{Bidirectional Depth-Augmented PnP (BD-PnP)} The BD-PnP layer converts the predicted revisions $\mathbf{r}$ and confidences $\mathbf{w}$ to a camera pose update $\Delta \mathbf{G}_0$. We first use the revisions to update the correspondence fields
\begin{equation}
\begin{split}
    \mathbf{x}_{i\rightarrow 0}' =  \mathbf{x}_{i\rightarrow 0} + \mathbf{r}_{i\rightarrow 0} \\
    \mathbf{x}_{0\rightarrow i}' =  \mathbf{x}_{0\rightarrow i} + \mathbf{r}_{0\rightarrow i}
\end{split}
\end{equation}
and define an objective function to minimize the distance between the reprojected coordinates and the revised correspondence 
\begin{equation}
\begin{split}
\label{eq:obj}
    \mathbf{E}(\mathbf{G}_0) = \sum_{i=1}^N \left||\mathbf{x}_{i\rightarrow 0}' - \Pi (\mathbf{G}_0\mathbf{G}_i^{-1}\Pi^{-1}(\mathbf{x}_i)|\right|_{\Sigma_{i\rightarrow0}}^2 + \\
    \sum_{i=1}^N \left||\mathbf{x}_{0\rightarrow i}' - \Pi (\mathbf{G}_i\mathbf{G}_0^{-1}\Pi^{-1}(\mathbf{x}_0)|\right|_{\Sigma_{0\rightarrow i}}^2
\end{split}
\end{equation}
where $||\cdot||_{\Sigma}$ is the Mahalanobis distance with $\Sigma_{i\rightarrow 0} = \text{diag} \ \mathbf{w}_{i \rightarrow 0}$. The objective in Eqn. \ref{eq:obj} states that we want camera poses $\mathbf{G}_0$ such that the reprojected points match the revised correspondence $\mathbf{x}_{ij}'$. It is important to note that this objective is similar to conventional PnP because it optimizes reprojection error. But unlike conventional PnP, which optimizes a single set of 2D-3D correspondences, our objective is bidirectional because it optimizes two sets of 2D-3D correspondences, one defined on the render and the other defined on the input image. In addition, unlike conventional PnP, our objective also includes reprojection error of inverse depth, which experiments show to be important for improving accuracy. 

We linearize Eqn.~\ref{eq:obj} using the current pose and perform a fixed number of Gauss-Netwon updates (3 during training and 10 during inference). Each Gauss-Newton update produces a pose update $\delta \xi \in \mathfrak{se}(3)$ which is applied to the current pose estimate using retraction on the SE3 manifold
\begin{equation}
    \mathbf{G}_0^{(t+1)} = \exp(\delta \xi) \cdot \mathbf{G}_0^{(t)}.
\end{equation}

\smallskip\noindent\textbf{Inner and Outer Update Loops}
For a given set of renders, we run 40 iterations of the update operator. Upon completion, we use the refined pose estimate to re-render a new set of 7 viewpoints and repeat the process. As we show in our experiments, we can trade speed for accuracy by increasing the number of inner and outer iterations.
\label{sec:loops}

\subsection{RGB Input} \label{sec:rgb-only} 

To handle RGB input, we can use the current pose $\mathbf{G}_0^{(t)}$ to render the depth from the known 3D model, and proceed as if we have RGB-D input. However, this basic approach is not mathematically sound because the rendered depth is a function of the object pose but is treated as a constant in the optimization. On the other hand, a fully principled treatment is difficult to implement because it requires computing the Jacobian of the rendering function, as well as the derivatives of the Jacobian during backpropagation. As a middle ground, we use the rendered depth to linearize the optimization objective, and introduce depth as a variable in the optimization so that we jointly optimize pose and depth but discard the depth update (full details in the appendix). This revised approach gives better results.

\subsection{Training}
Each training step, we randomly sample a visible object in the input image and randomly purturb the ground-truth rotation and translation to initialize the pose. Our model is trained to recover the ground truth pose from this initialization. In order to save GPU memory, we use 10 inner-loops and one outer-loop during training, and render only one viewpoint at each training step.

 \smallskip \noindent
\textbf{Supervision} We supervise on the predicted correspondence revisions and the updated pose estimates from all update iterations in the forward pass, with exponentially increasing loss weights similar to RAFT~\cite{teed2020raft}. Specifically, we supervise on the geodesic L1 distance between the estimated pose and the ground truth pose. The flow is supervised using an L1 endpoint error loss, as is standard for optical flow problems. All ground truth poses in the BOP benchmark\cite{bopchallenge}, which we use for experiments, have a set of discretized symmetries which are considered equivalent with regard to the MSSD and MSPD error metrics. In order to align the loss with the error metrics, we compute the loss using all discretized symmetries and backpropagate the minimum.

\section{Experiments}
\begin{table}[]
\centering
\resizebox{.5\textwidth}{!}{
\begin{tabular}{lcccc}
\toprule
Method & Avg. & MSPD & VSD & MSSD \\
\midrule
\multicolumn{5}{c}{\underline{YCB-V}\cite{ycb}}\\

\midrule
\textbf{Ours}&\textbf{0.893}&\textbf{0.885}&\textbf{0.871}&\textbf{0.924}\\
CosyPose\cite{cosypose}&\underline{0.861}&	\underline{0.849}&\underline{0.831}&\underline{0.903}\\
W-PoseNet w/ICP \cite{posenet}&0.779&0.734&0.779&0.824\\
Pix2Pose-BOP20 (w/ ICP) \cite{pix2pose}&0.780&	0.758&0.766&0.817\\
Koenig-Hybrid-DL-PointPairs \cite{koenig} &0.701&	0.635&0.778&0.690\\
CDPNv2-BOP20 (w/ ICP) \cite{cdpn}&0.619&	0.565&0.590&0.701\\
EPOS \cite{epos} & 0.696&0.783&	0.626&0.677\\
Vidal-Sensors18 \cite{vidal} &0.450&0.347&0.623&0.380\\
\midrule
\multicolumn{5}{c}{\underline{T-LESS \cite{tless}}}\\

\midrule
\textbf{Ours}&\textbf{0.776}&\underline{0.795}&\textbf{0.760}&\textbf{0.773}\\
CosyPose \cite{cosypose}& \underline{0.728}&\textbf{0.821}&\underline{0.669}&\underline{0.695}\\

Pix2Pose-BOP20 (w/ ICP) \cite{pix2pose} & 0.512&0.549&0.438&0.548\\
Koenig-Hybrid-DL-PointPairs \cite{koenig} &0.655&0.696&	0.580&0.689\\
CDPNv2-BOP19 (w/ ICP) \cite{cdpn}&0.490&	0.674&0.377&0.418\\
EPOS\cite{epos} &0.476&	0.635&0.369&0.423\\
Vidal-Sensors18 \cite{vidal}& 0.538&	0.574&0.464&0.575\\
\midrule
\multicolumn{5}{c}{\underline{LINEMOD-Occluded}\cite{linemod_occlusion}}\\

\midrule
\textbf{Ours}&\textbf{0.734}&\underline{0.824}&\textbf{0.601}&\textbf{0.778}\\
CosyPose \cite{cosypose}& \underline{0.714}&\textbf{0.826}&0.567&\underline{0.748}\\
W-PoseNet w/ICP \cite{posenet} &0.707&0.793&\textbf{0.601}&0.726\\
Pix2Pose-BOP20  (w/ ICP) \cite{pix2pose}&0.588&	0.659&0.473&0.631\\
Koenig-Hybrid-DL-PointPairs \cite{koenig}  & 0.631&0.703&0.517&0.675\\
CDPNv2-BOP20 (w/ ICP) \cite{cdpn}&0.630&0.731&0.469&0.689\\
EPOS \cite{epos} &0.547&	0.750&0.389&0.501\\
Vidal-Sensors18 \cite{vidal} & 0.582&	0.647&	0.473&0.625\\
\bottomrule
\end{tabular}
}
\caption{Top performing methods on the BOP Benchmark \cite{bopchallenge}. The MSPD, VSD and MSSD columns are their recall across a range of thresholds (sec. \ref{sec:eval_metrics}). We use the same detector as cosypose \cite{cosypose}}
\label{tab:rgbd_final}
\end{table}

\begin{table}[h]
\centering
\resizebox{.85\columnwidth}{!}{
\begin{tabular}{lcccc}
\toprule
Method & Avg. & MSPD & VSD & MSSD \\
\midrule
\multicolumn{5}{c}{\underline{YCB-V \cite{ycb}}}\\

\midrule
\textbf{Ours}&\textbf{0.824}&\textbf{0.852}&\textbf{0.783}&\underline{0.835}\\
CosyPose \cite{cosypose}&\underline{0.821}&\underline{0.850}&	\underline{0.772}&\textbf{0.842}\\
EPOS \cite{epos}&0.696&	0.783&0.626&0.677\\
CDPN \cite{cdpn} & 0.532&0.631&	0.396&0.570\\
\midrule
\multicolumn{5}{c}{\underline{T-LESS \cite{tless}}}\\

\midrule
\textbf{Ours}&\underline{0.715}&\underline{0.798}&\underline{0.663}&\underline{0.684}\\
CosyPose \cite{cosypose}&\textbf{0.728}&\textbf{0.821}&\textbf{0.669}&\textbf{0.695}\\
EPOS \cite{epos}&0.476&0.635&0.369&0.423\\
CDPN \cite{cdpn}&0.490 & 	0.674 & 0.377 & 0.418\\
\midrule
\multicolumn{5}{c}{\underline{LINEMOD-Occluded \cite{linemod_occlusion}}}\\

\midrule
\textbf{Ours}&\textbf{0.655}&\textbf{0.831}&\textbf{0.501}&\textbf{0.633}\\
CosyPose \cite{cosypose}&\underline{0.633}&	\underline{0.812}&	\underline{0.480}&0.606\\
EPOS \cite{epos}&0.547&0.750&0.389&0.501\\
CDPN \cite{cdpn}&0.624&0.731&	0.469&\underline{0.612}\\
\bottomrule
\end{tabular}
}
\caption{Results on the BOP Benchmark \cite{bopchallenge} excluding methods that use depth. The MSPD, VSD and MSSD columns are their recall across a range of thresholds (sec. \ref{sec:eval_metrics}). We use the same detector as cosypose \cite{cosypose}.\vspace{-0.4cm}
}
\label{tab:rgb_only_final}
\end{table}

\label{sec:eval_metrics}
\noindent\textbf{Evaluation Metrics} In keeping with the BOP benchmark evaluation, we report Maximum Symmetry-Aware Surface Distance (MSSD) Recall, Maximum Symmetry-Aware Projection Distance (MSPD) Recall, Visible Surface Discrepancy (VSD) Recall, as well as the average over all three metrics. MSSD is the maximum euclidean distance between associated mesh vertices in the predicted and ground truth poses. MSPD is the maximum reprojection error between all associated vertices from the predicted and ground truth poses. Both MSSD and MSPD assume the minimum value across all symmetrically equivalent ground truth poses. VSD is the depth discrepancy between the mesh rendered at the predicted and ground truth poses, measured over pixels where the model is visible. All three metrics are reported as recall percentages over a set of thresholds defined in the BOP benchmark \cite{bopchallenge}, scaled between 0 and 1. Higher is better for all three.  

\smallskip \noindent\textbf{Datasets} We evaluate our method on the \textit{varying number of instances of a varying number of objects} in a single RGB-D image (the ViVo task) from the official BOP benchmark \cite{bopchallenge}. Specifically, we evaluate our method on the YCB-V \cite{ycb}, T-LESS \cite{tless} and LM-O (Linemod-Occluded) \cite{linemod_occlusion} datasets from the BOP benchmark \cite{bopchallenge}. Each dataset consists of a unique set of objects designed to evaluate a method's accuracy in different real-world settings. The YCB-V dataset consists of 21 household objects with texture and color, the T-LESS dataset consists of 30 highly-similar industry-relevant objects with no texture or color, and the Linemod (Occluded) dataset consists of 15 texture-less colored household objects. For each image in the test sets, we must classify and predict the rotation and translation of all visible objects. There are 900 YCB-V test images, 1000 T-LESS test images and the 200 Linemod-Occluded test images. Each image contains 3 to 8 objects. 

\smallskip\noindent\textbf{Training Data} On the YCB-V Objects dataset \cite{ycb}, we use the 80K synthetic and 113K real training images provided in the BOP challenge \cite{bopchallenge}. On the T-LESS \cite{tless} dataset, we train using the 50K synthetic and 38K real training images. On the Linemod dataset, we train exclusively using the 50K synthetic training images provided. Please see the appendix for additional implementation details. 

\subsection{BOP Benchmark Results}
\begin{table*}
\centering
\resizebox{\linewidth}{!}{
\begin{tabular}{lcccccc}
\toprule
&\multicolumn{2}{c}{YCB-V \cite{ycb}}&\multicolumn{2}{c}{T-LESS \cite{tless}} &\multicolumn{2}{c}{LM-O \cite{linemod_occlusion}} \\
\cmidrule(lr){2-3}
\cmidrule(lr){4-5}
\cmidrule(lr){6-7}
&MSPD Recall&MSSD Recall&MSPD Recall&MSSD Recall&MSPD Recall&MSSD Recall \\
\midrule
\textbf{Bidirectional PnP}&\textbf{0.924}&\textbf{0.955}&\textbf{0.685}&\textbf{0.582} & \textbf{0.828} &\textbf{0.788} \\
Unidirectional PnP (render to image)&0.905&0.941&0.677&0.546& 0.605 & 0.465 \\
Unidirectional PnP (image to render)&0.890&0.917&0.337&0.200& 0.811 & 0.773 \\
\midrule
\textbf{Depth-Augmented PnP (predicting depth revisions)} &\textbf{0.924}&\textbf{0.955}&\textbf{0.685}&\textbf{0.582}&\textbf{0.828} &\textbf{0.788} \\
No depth augmentation (no depth revisions) & 0.909&0.940&0.678&0.573&0.819&0.784 \\

\midrule
\textbf{Predicting per-pixel confidence weights}&\textbf{0.924}&\textbf{0.955}&\textbf{0.685}&\textbf{0.582} & \textbf{0.828} &\textbf{0.788} \\
Uniform confidence &0.721&0.832&0.587&0.424&0.812	& 0.760 \\
\midrule
\textbf{Multiview renders}&\textbf{0.924}&\textbf{0.955}&\textbf{0.685}&\textbf{0.582} & \textbf{0.828} &\textbf{0.788} \\
Single render&{0.902}&{0.941}&{0.663}&{0.545}&0.744	& 0.641 \\
\midrule
\textbf{Coupled iterative refinement} &\textbf{0.924} & \textbf{0.955} & \textbf{0.685} & \textbf{0.582}& \textbf{0.828} &\textbf{0.788} \\
One-shot (flow by RAFT  followed by PnP) &0.562&0.643&0.483&0.275&0.569&0.054 \\
\midrule
\textbf{Pose + Flow Loss}&\textbf{0.924}&\textbf{0.955}&\textbf{0.685}&\textbf{0.582} & \textbf{0.828} &\textbf{0.788} \\
Flow Loss Only&0.740&0.733&0.558&0.386&0.804&0.735 \\
Pose Loss Only&0.866&0.919&0.261&0.169&0.615&0.303 \\
\midrule

\textbf{4 Outer Loops}&\textbf{0.933}&\textbf{0.958}&\textbf{0.694}&\textbf{0.601}&\textbf{0.831}&0.787 \\
1 Outer Loop&0.924&0.955&0.685&0.582&0.828 &\textbf{0.788} \\

\midrule
No refinement of initial pose & 0.194&0.298&0.263&0.167&0.475&0.316 \\
\bottomrule
\end{tabular}
}
\caption{Ablation experiments using our method for RGB-D input. We evaluate our method on held-out training images. Initial poses are generated by randomly perturbing the ground truth pose.  Options used in our full method are bolded. }
\label{tab:ablations_rgbd}
\end{table*}

\noindent \textbf{RGB-D Results} Our approach significantly outperforms all other methods on YCB-V, T-LESS and LM-O for RGB-D input (see Tab. \ref{tab:rgbd_final}). Inline with the BOP benchmark guidelines, all of our RGB-D methods use the exact same hyper parameter settings across all datasets. For each prior work against which we compare, we report their best single-image method per-dataset, with or without ICP, whichever is better. Our reported results do not use ICP.

\noindent\textbf{RGB-Only Results} Using our adaptation to RGB-only input, we compare our method to all prior work on the BOP benchmark. Our method performs competitively with the state-of-the-art on the BOP benchmark (See Tab. \ref{tab:rgb_only_final}), outperforming on Linemod (Occluded) and YCB-V, and underperforming on T-LESS. Just as with our RGB-D results, all our RGB results were obtained using the same hyper parameter settings.

\subsection{Ablation Experiments}

\begin{table*}[]
    \centering
    \resizebox{\textwidth}{!}{
    \begin{tabular}{lcccccc}
    \toprule
&\multicolumn{2}{c}{YCB-V \cite{ycb}}&\multicolumn{2}{c}{T-LESS \cite{tless}}&\multicolumn{2}{c}{LM-O\cite{linemod_occlusion}} \\
\cmidrule(lr){2-3}
\cmidrule(lr){4-5}
\cmidrule(lr){6-7}
&MSPD Recall&MSSD Recall&MSPD Recall&MSSD Recall&MSPD Recall&MSSD Recall \\

\midrule
\textbf{Revised approach (depth as variable)} & \textbf{0.833}&\textbf{0.751}&\textbf{0.649}&\textbf{0.474}&0.793&\textbf{0.609} \\
Basic approach (depth as constant) &0.814&0.678&0.637&0.343&\textbf{0.808}&0.570 \\

\bottomrule
    \end{tabular}}
    \caption{Ablation experiments using our method for RGB-Only input. We evaluate our method on a held-out subset of training images. Initial poses are generated by randomly perturbing the ground truth pose. Options used in our full RGB method are bolded. Additional ablations on our RGB method are in the supplemenary material.
    }
    \label{tab:ablations_rgb}
\end{table*}

All ablation experiments were conducted on held-out scenes from the training data in YCB-V, T-LESS, and LM-O. We mainly report ablation results for the RGB-D setting (Tab.~\ref{tab:ablations_rgbd}). We perform  the same ablations for the RGB setting and report the results in the appendix. The better design choices based on RGB-D ablations also perform well for RGB input, and are used to obtain our final results for both settings. 

\smallskip
\noindent \textbf{Bidirectional Depth-Augmented PnP} Our method benefits from being bidirectional between an image-render pair. Using correspondence from only a single direction in the PnP solver yields less accurate results. In addition, depth augmentation improves accuracy.

\smallskip
\noindent\textbf{Predicting Confidence Weights} The predicted confidence weights allow our model to down-weight correspondences which are outliers. This behavior is critical to performance, as uniform confidence over all pixels within the masks is much less accurate (see Tab. \ref{tab:ablations_rgbd}). 

\smallskip
\noindent\textbf{Multi-view Renders} In the forward pass, we can arbitrarily add more viewpoints by rendering the object at additional perturbations of the input pose. The memory usage scales linearly with the number of viewpoints, therefore during inference this is tractable. Even without explicitly training using more than one render, rendering seven rotationally-perturbed viewpoints leads to better results across the board during inference (see Tab. ~\ref{tab:ablations_rgbd}). Viewpoints generated from excessively large rotation perturbations share too few correspondences with the input image to be useful, while too small perturbations add little novel information. $22.5^{\circ}$ perturbations work well in all situations.

\smallskip
\noindent\textbf{Coupled Iterative Refinement} Tightly coupled iterative refinement of both correspondence and pose performs substantially better than the single-shot approach that solves for pose after predicting correspondence.

\smallskip
\noindent \textbf{Flow and Pose Loss} Both the pose and flow loss functions are critical, suggesting that there is strong coupling between the pose updates and flow updates. 

\smallskip
\noindent \textbf{Outer Loop} It is beneficial to have an outer loop that regenerates the renders with the latest pose estimate. 

\smallskip \noindent \textbf{Handling RGB-only Input} Tab.~\ref{tab:ablations_rgb} compares the basic approach and the revised approach for handling RGB-only input (see Sec.~\ref{sec:rgb-only}). The revised approach performs better.

\subsection{Speed vs Accuracy Tradeoff}

\begin{figure}[t]
    \centering
	\includegraphics[width=1.0\linewidth]{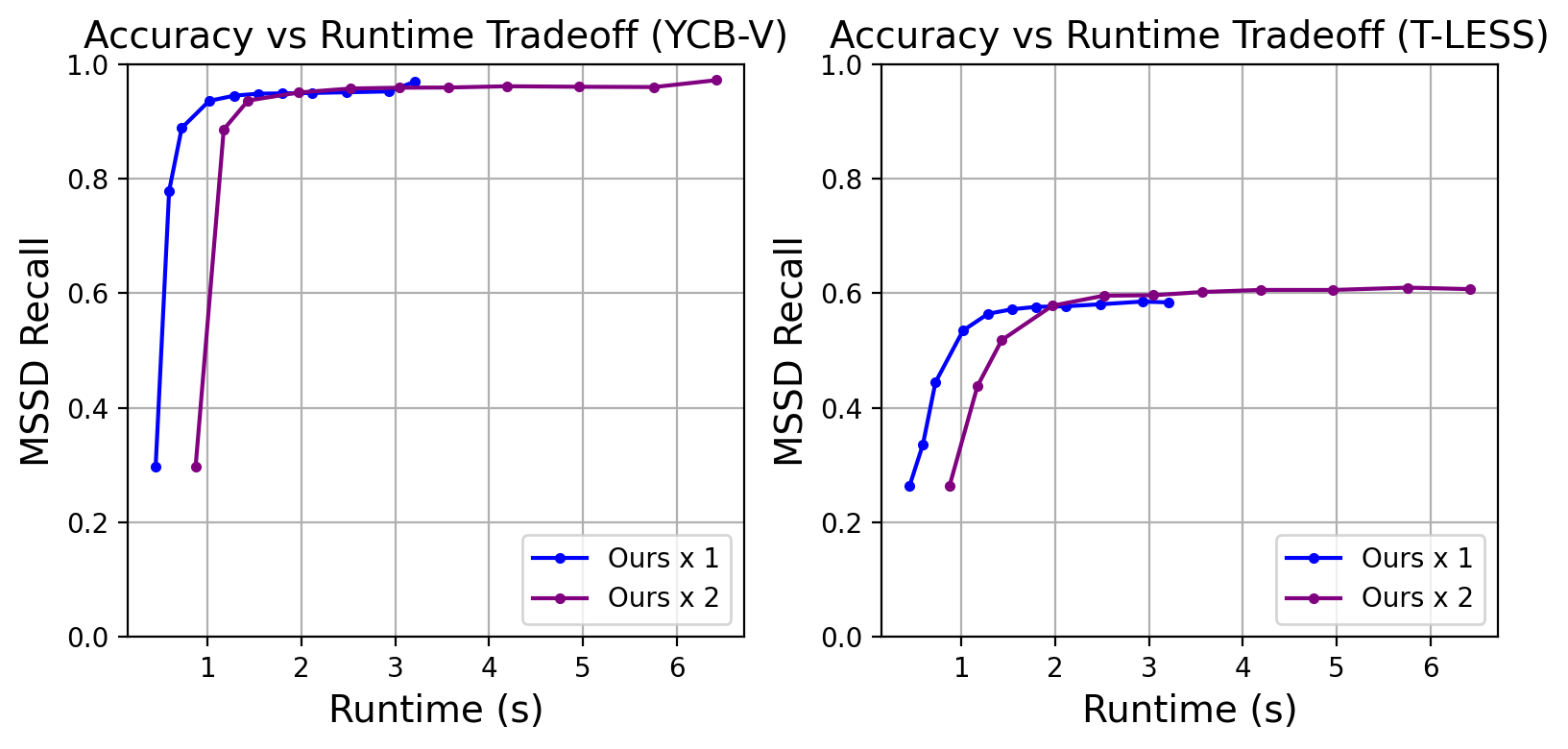}
	\caption{The accuracy / speed tradeoff of our method. Results of one outer update loop (Ours 1x) and two (Ours 2x). \textbf{Left:} Accuracy on a held-out split of YCB-V training data. \textbf{Right:} Accuracy on a held-out split of T-LESS training data. Our method converges quickly, meaning few inner loops and one outer loop gives good results. Timings are measured on single objects with random rotation and translation perturbation.}
	\label{fig:runtime}
\end{figure}
One can trade off accuracy for speed by varying the number of outer or inner update loops (sec. \ref{sec:loops}). In Fig. \ref{fig:runtime}, we report the accuracy of our method on a held-out portion of the YCB-V and T-LESS training datasets as a function of runtime, for both one and two outer-loops. For our final results in Tab. \ref{tab:rgbd_final} and \ref{tab:rgb_only_final}, we use 4 outer loops and 40 inner loops, which takes 10.80s per batch of detections. However, Fig. \ref{fig:runtime} shows that our method converges quickly with few inner loops and one outer loop. 
\subsection{Robustness}
\begin{figure}[h]
    \centering
	\includegraphics[width=1.0\linewidth]{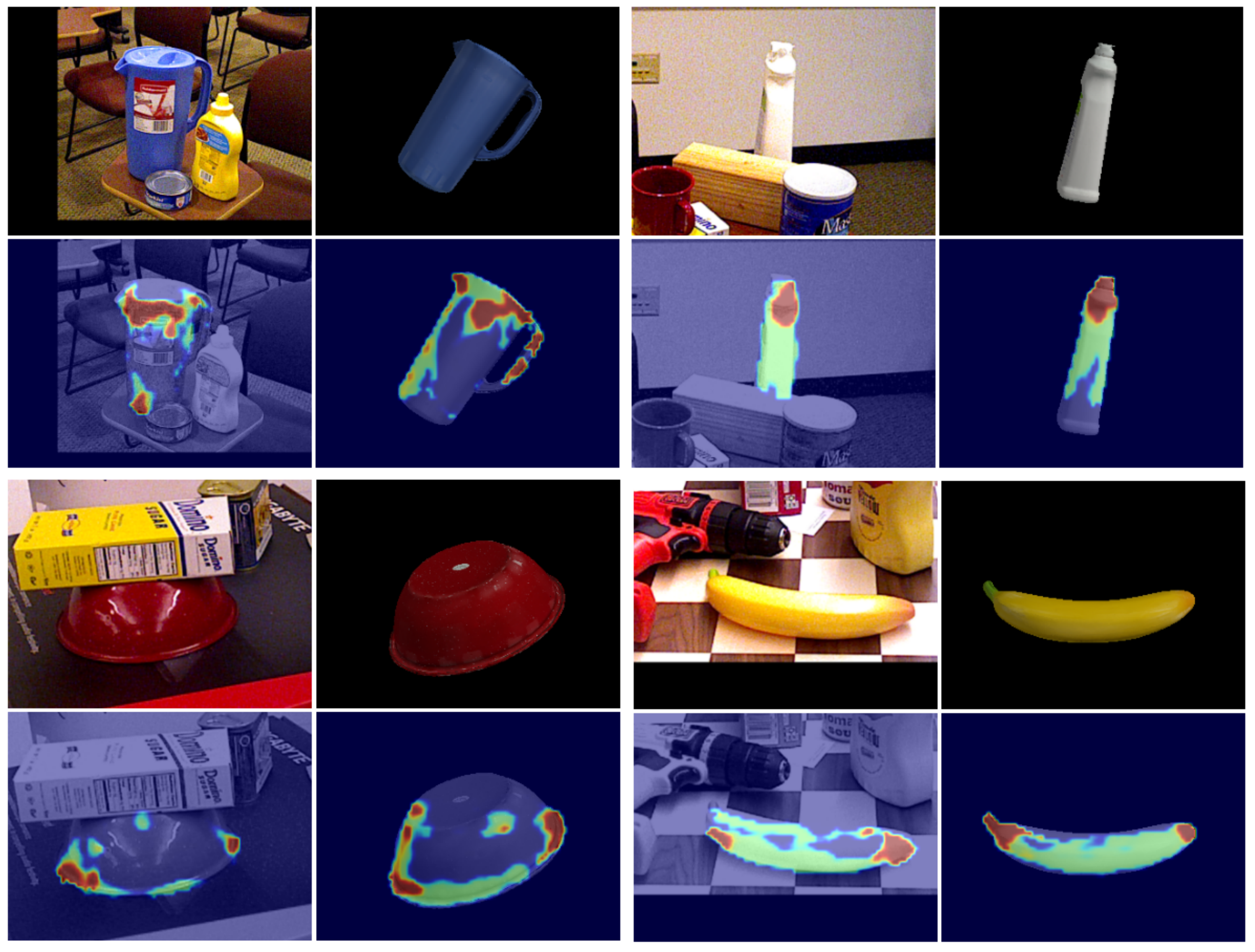}
	\caption{The predicted confidence weights on the YCB-V dataset. The heatmaps provide insight into what surface features are most helpful for pose optimization algorithms. Specifically, our method has low confidence over textureless regions, and high confidence over textured ones, over thin structures, and on edges.}\vspace{-0.5cm}
	\label{fig:heatmaps_ycbv}
\end{figure}
\begin{figure}[t]
    \centering
	\includegraphics[width=1.0\linewidth]{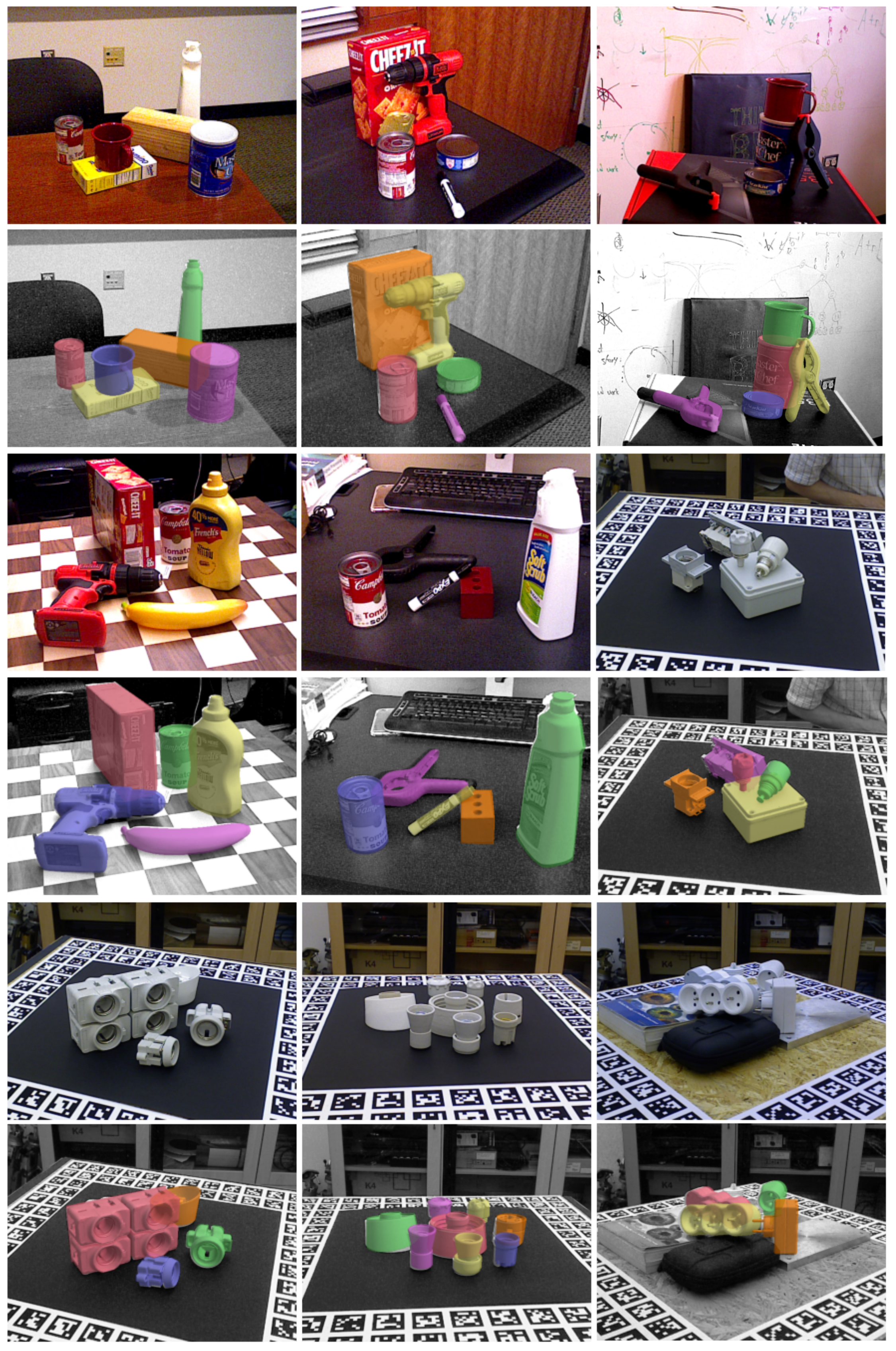}
	\caption{Predictions on the YCB-V and T-LESS test datasets. Known object models are rendered at the predicted poses. }
	\label{fig:qual_results}
\end{figure}

\begin{figure}[t]
    \centering
	\includegraphics[width=1.0\linewidth]{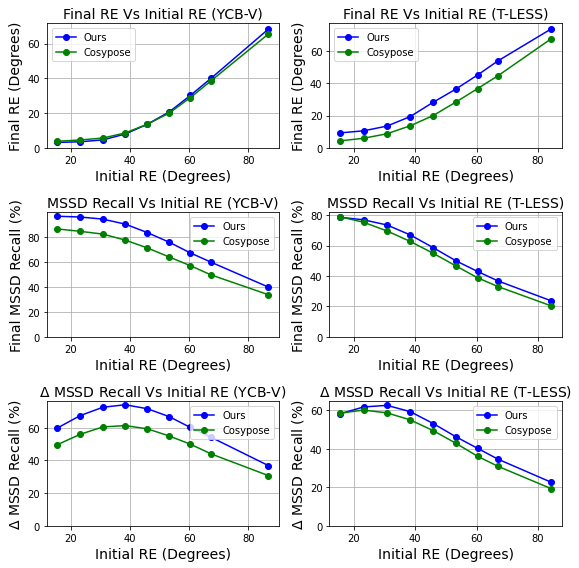}
	\caption{Our model is robust to partially incorrect initial input poses. We plot the accuracy (MSSD recall) and rotation error of the output pose as a function of the rotation error (SO3 geodesic distance, denoted ``RE'') of the initial input pose. Input poses were randomly rotated from the ground-truth. \textbf{Top:} Output rotation error. Our RE is slightly higher on T-LESS. \textbf{Middle:} Output MSSD Recall. Our method is more accurate. \textbf{Bottom:} Improvement to MSSD Recall over the MSSD Recall of the initial pose.}
	\label{fig:starting_pose}
\end{figure}
We evaluate the robustness of our method to inaccurate initial pose estimates on the YCB-V test set. In addition to a coarse pose estimation model, Cosypose~\cite{cosypose} introduces a regression-based refinement model. In Fig.~\ref{fig:starting_pose}, we plot the accuracy of our model as a function of the rotation error of the initial input pose. For comparison, we also include the refinement model introduced in \cite{cosypose}. Both methods were trained using the same random perturbations of the training data. A limitation of our method is that its ability to refine the poses diminishes for larger initial rotation errors.
\section{Qualitative Results}

\begin{figure}[t]
    \centering
	\includegraphics[width=1.0\linewidth]{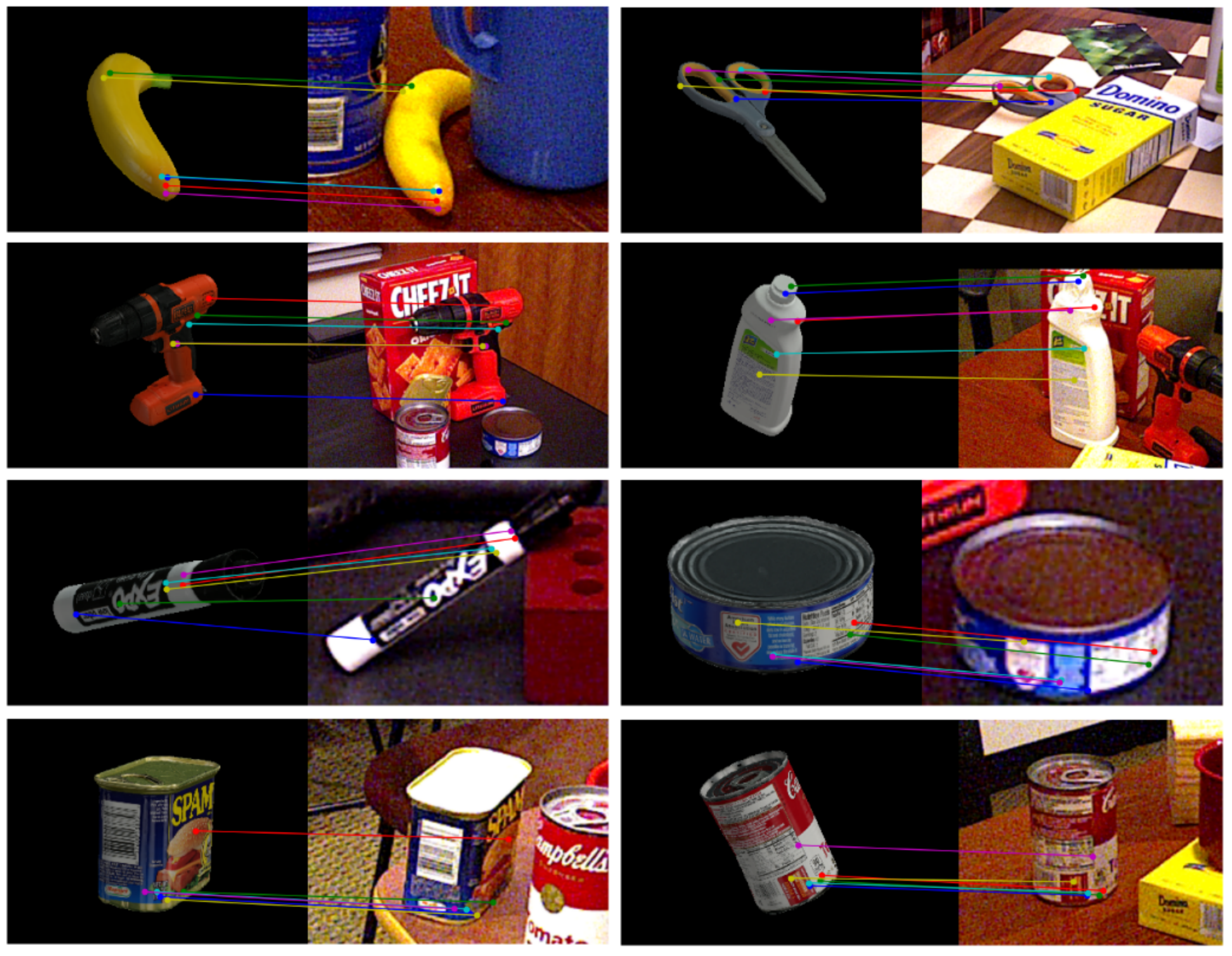}
	\caption{The predicted high-confidence matches on the YCB-V dataset between the input image and the rendered input pose. We apply non-max suppression with 5-pixel radius to the confidence weights and show the most confident predicted correspondences. Our method learns to predict high confidence for matches that are useful for solving for pose.\vspace{-0.2cm}}
	\label{fig:matches_ycbv}
\end{figure}

\smallskip \noindent \textbf{Confidence Weights} In the forward pass, our model generates a dense field of confidence weights for all predicted correspondences between an image and rendered pose estimate. In Fig. \ref{fig:heatmaps_ycbv}, we visualize these confidence weights as heatmaps over the images and renders. The heatmaps indicate which parts of the images are most useful for predicting the object's pose. In Fig. \ref{fig:matches_ycbv}, we show the correspondences with highest confidence within 5-pixel radii. 

\smallskip 
\noindent \textbf{Full Image Predictions} In Fig. \ref{fig:qual_results}, we show the results of our end-to-end method for multi-object pose prediction on the T-LESS and YCB-V test datasets. Additional qualitative results on the YCB-V, T-LESS and LM-O test datasets are included in the appendix.
\section{Conclusion} We have introduced a new approach to 6D multi-object pose estimation. Our approach iteratively refines both pose and dense correspondence together using a novel differentiable solver layer. We also introduce a variant of our method for RGB-only input. Our method achieves state-of-the-art accuracy on standard benchmarks.

\noindent \textbf{Acknowledgements} This work is partially supported by the National Science Foundation under Award IIS-1942981.

{\small
\bibliographystyle{ieee_fullname}
\bibliography{egbib}
}

\appendix

\onecolumn

\begin{center}
    \Large \textbf{Coupled Iterative Refinement for 6D Multi-Object Pose Estimation Appendix}
\end{center}

\begin{figure}[b]%
    \centering
    \subfloat[\centering RGB Pose Update Module]{{\includegraphics[width=0.40\textwidth]{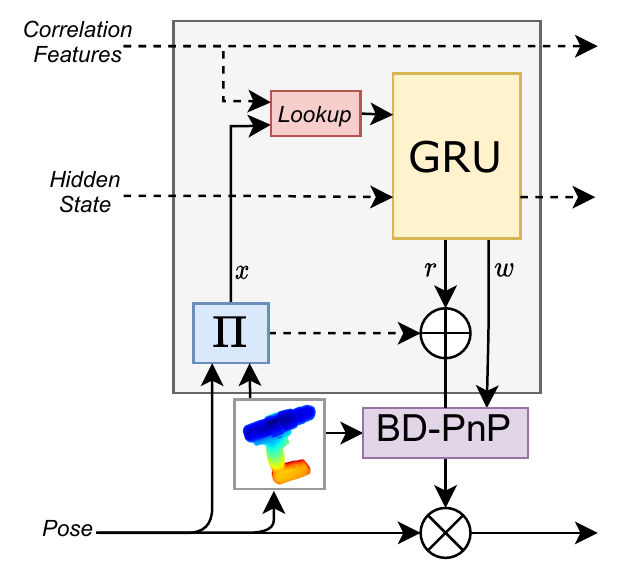}\label{pose_update_rgb}}}
    \qquad
    \subfloat[\centering Context features and hidden state initialization]{{\includegraphics[width=0.50\textwidth]{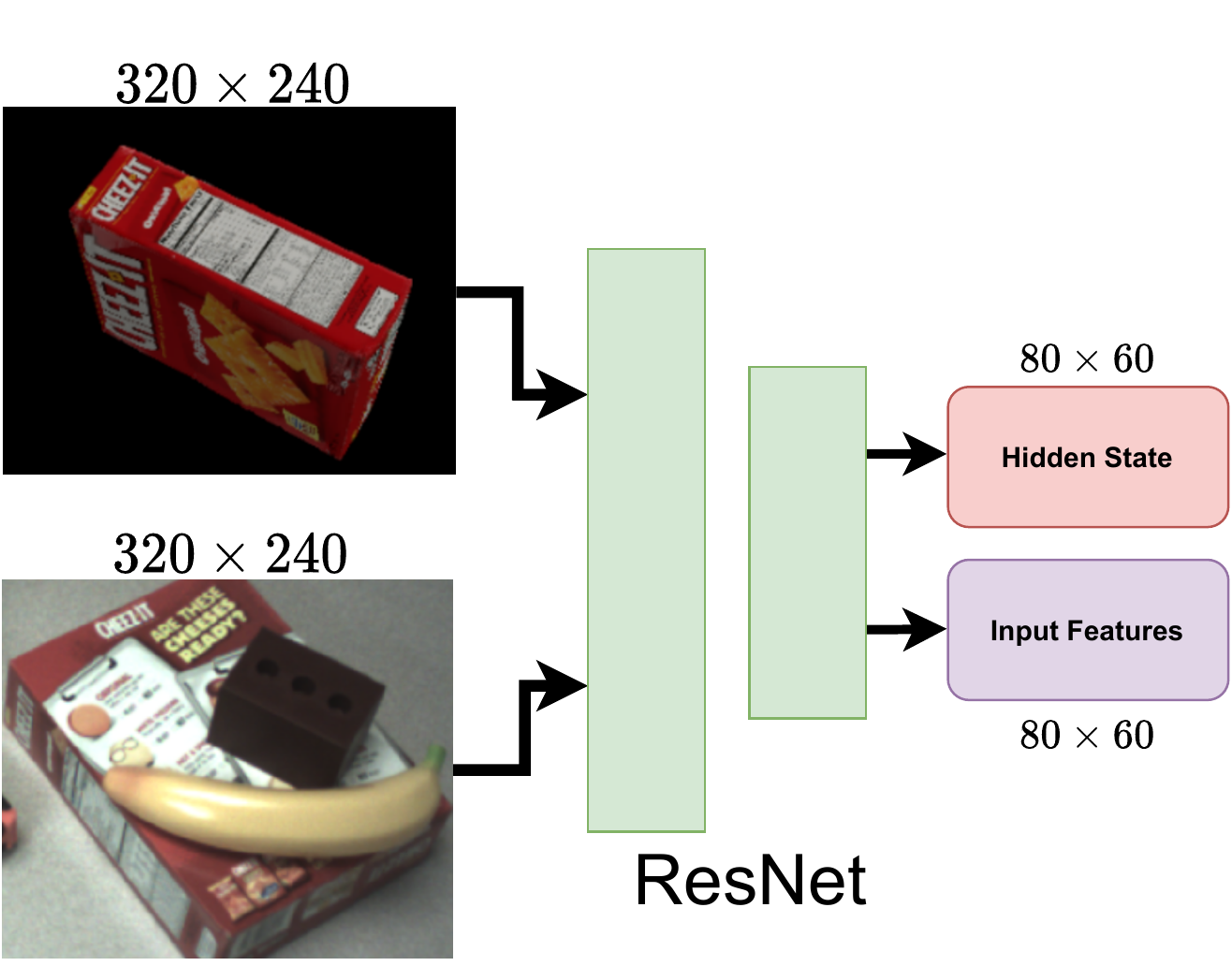}\label{cont_encoders}}}%
\end{figure}

The correlation features and hidden state input to the GRU are described in the main paper. The additional inputs to the GRU are described in this section.

\smallskip
\noindent\textbf{Context Features} 
Following the procedure described in RAFT \cite{teed2020raft}, we use a Resnet\cite{resnet}-based feature extractor to construct context features and an initial hidden state for every image (Fig. \ref{cont_encoders}). The hidden state is updated with every GRU iteration, while the context features remain unchanged.

\smallskip
\noindent\textbf{Depth Features}
For each image pair $(i,0)$, we have inverse-depth maps $\mathbf{z}^{-1}_i$ and $\mathbf{z}^{-1}_0$. We use $\mathbf{x}_{i\rightarrow0}$ to index $\mathbf{z}^{-1}_0$, producing a new inverse-depth map $\mathbf{z}^{-1}_{i\rightarrow0}$. The depth residuals $(\mathbf{z}^{-1}_0 - \mathbf{z}^{-1}_{i\rightarrow0})$ implied by the induced correspondences $\mathbf{x}_{i\rightarrow0}$ are provided as input to the GRU before each update. This gives the GRU more information on how well depth has been aligned. Since $\mathbf{x}_{i\rightarrow0}$ are real numbers, we use bilinear interpolation. This procedure is identical for correspondences from image $0$ to $i$. 

\smallskip
\noindent\textbf{Solver Residuals}
The solver residuals from the previous solver iteration are fed to the GRU. These residuals are calculated by taking the difference between the current induced correspondences $\mathbf{x}^{(t)}_{i\rightarrow0}$ and the previous revised correspondences $\mathbf{x}^{\prime(t-1)}_{i\rightarrow0}$. This allows our model to detect outliers easily as the pixels with unusually high residuals, and thus prune them in the next update. 

\section{RGB-Only Optimization}
\label{sec:rgb_optim}
In the RGB-D setting, depth for the input image is given and stays fixed for each pixel regardless of the current pose estimate. In the RGB-Only setting, depth is obtained by rendering the object at the current pose estimate $\mathbf{G}_0$. See Fig. \ref{pose_update_rgb}.
\begin{equation}
    \mathbf{x}_0^{RGBD} = \begin{bmatrix} x_0 \\ y_0 \\ d_0 \end{bmatrix} \ \ \mathbf{x}_0^{RGB} = \begin{bmatrix} x_0 \\ y_0 \\ \mathcal{R}(\mathbf{G}_0) \end{bmatrix}
\end{equation}
\noindent In our objective function, $\mathbf{G}_0$ maps the points between images. In the RGB setting, it also produces the depth for points $\mathbf{x}^{RGB}_0$ in the input image. 

\begin{equation}
\begin{split}
\label{eq:obj_rgbd}
    \textbf{RGBD}:\qquad \mathbf{E}(\mathbf{G}_0) = \sum_{i=1}^N \left||\mathbf{x}_{i\rightarrow 0}' - \Pi (\mathbf{G}_0\mathbf{G}_i^{-1}\Pi^{-1}(\mathbf{x}_i))|\right|_{\Sigma_{i\rightarrow0}}^2 + \sum_{i=1}^N \left||\mathbf{x}_{0\rightarrow i}' - \Pi (\mathbf{G}_i\mathbf{G}_0^{-1}\Pi^{-1}(\mathbf{x}^{RGBD}_0))|\right|_{\Sigma_{0\rightarrow i}}^2
\end{split}
\end{equation}
\begin{equation}
\begin{split}
\label{eq:obj_rgb}
    \textbf{RGB}:\qquad \mathbf{E}(\mathbf{G}_0) = \sum_{i=1}^N \left||\mathbf{x}_{i\rightarrow 0}' - \Pi (\mathbf{G}_0\mathbf{G}_i^{-1}\Pi^{-1}(\mathbf{x}_i))|\right|_{\Sigma_{i\rightarrow0}}^2 + \sum_{i=1}^N \left||\mathbf{x}_{0\rightarrow i}' - \Pi (\mathbf{G}_i\mathbf{G}_0^{-1}\Pi^{-1}(\mathbf{x}^{RGB}_0))|\right|_{\Sigma_{0\rightarrow i}}^2
\end{split}
\end{equation}
In the RGB setting, the depth of $\mathbf{x}^{RGB}_0$ is a function of the pose $\mathbf{G}_0$ in Eq. \ref{eq:obj_rgb}. However, it is difficult to treat it as such in the optimization since our differentiable solver requires calculating the derivative of the Jacobian for the rendering function $\mathcal{R}$. Instead, we treat the depth in the image as a variable to be optimized over jointly with the pose $\mathbf{G}_0$. The resulting optimized pose and depth may be inconsistent with one another since they were treated as separate variables in the optimization. Therefore, we discard the depth update produced by the solver and produce a new depth map by rendering the updated pose. This ensures that the depth is a function of the pose.

\section{Implementation Details:}

\noindent\textbf{Training Schedule:} Our method is implemented in Pytorch \cite{pytorch}. All models are initialized from scratch with random weights and trained for 200K steps. During training, we use the AdamW \cite{adamw} optimizer. Final models are trained with a batch size of 12 on two RTX-3090s. Ablation experiments are trained with a batch size of 4. We use an exponential learning rate schedule with a linear increase to $3\times 10^{-4}$ over $10000$ steps and a $50\%$ drop every subsequent $20000$ steps. We use $10^{-5}$ weight decay. We use the full resolution images for training: $640\times 480$ for YCB-V, $720\times 540$ for T-LESS, and $640\times 480$ for LM-O. In the inner update loop, the correspondence field, confidence weights, depth, hidden state, etc. are maintained at $80 \times 60$ spatial resolution, which is $\frac{1}{4}$ of the resolution of the $320 \times 240$ input crop. We downsample the input depth by subsampling and use strided convolutions for the image features.

\smallskip
\noindent\textbf{Image Augmentation:} We use the same image augmentation as \cite{cosypose} on all three datasets, specifically contrast, hue, sharpness, gaussian blur, and brightness. The depth images in the training data are sparse, so we fill in the gaps using bilinear interpolation. When training our RGB-Only variant, we also swap the ground truth segmentation mask with one predicted by a Mask R-CNN in order to prevent overfitting to the mask boundary.

\smallskip
\noindent\textbf{Multi-view Renders:} At test time, our RGB-D method renders 6 additional views at ${22.5}^{\circ}$ perturbations around the current pose estimate in the beginning of each outer-loop, for a total of 7. Our RGB-Only method renders 6 \textit{more} views at ${45}^{\circ}$ perturbations, for a total of 13.

\section{Accuracy Metrics} 

Our ablation experiments and main results follow the evaluation protocol used in the BOP Challenge \cite{bopchallenge}. Here, we formally define the error metrics used.

\smallskip
\noindent \textbf{Visible Surface Discrepancy (VSD)}
\begin{align*}
&e_\mathrm{VSD} =
\underset{p \in \hat{V} \cup \bar{V}}{\mathrm{avg}}
\begin{cases}
0 & \text{if} \, p \in \hat{V} \cap \bar{V} \, \wedge \, |\hat{D}(p) -
\bar{D}(p)| < \tau \\
1 & \text{otherwise},
\end{cases}
\end{align*}
where $\hat{D}(p)$ and $\bar{D}(p)$ are the depth maps obtained by rendering the object at the predicted pose and ground-truth pose, respectively. $\hat{V}$ and $\bar{V}$ are visibility masks obtained by comparing each depth map with the sensor depth. VSD treats indistinguishable poses as identical. VSD Recall is the percent of VSD scores less than $10$ thresholds ranging from $0.05$ to $0.5$, with the misalignment tolerance $\tau$ ranging from $5\%$ to $50\%$ of the object's diameter.

\smallskip
\noindent \textbf{Maximum Symmetry-Aware Surface Distance (MSSD)}
\begin{align*}
&e_{\text{MSSD}} = \text{min}_{\textbf{S} \in S_O} \text{max}_{\textbf{x}
\in V_O}
\Vert \hat{\textbf{P}}\textbf{x} - \bar{\textbf{P}}\textbf{S}\textbf{x}
\Vert_2,
\end{align*}
where $S_O$ are the rotation symmetries of object $O$ and $V_O$ are its vertices. $\hat{\textbf{P}}$ and $\bar{\textbf{P}}$ are the predicted and ground truth poses. MSSD is useful for robotic manipulation where the maximum surface deviation is related to the chance of a successful grasp. Compared to the average distance used in ADD/ADI, the maximum distance is less dependant on the sampling density of vertices. MSSD Recall is the percent of MSSD scores less than $10$ thresholds ranging from $5\%$ to $50\%$ of the object's diameter.

\smallskip
\noindent \textbf{Maximum Symmetry-Aware Projection Distance (MSPD)}
\begin{align*}
&e_{\text{MSPD}} = \text{min}_{\textbf{S} \in S_O} \text{max}_{\textbf{x}
\in V_O}
\Vert \text{proj}( \hat{\textbf{P}}\textbf{x} ) - \text{proj}(
\bar{\textbf{P}}\textbf{S}\textbf{x} ) \Vert_2,
\end{align*}
where $V_O$, $S_O$, $\hat{\textbf{P}}$, $\bar{\textbf{P}}$ are defined above. MSPD evaluates the perceivable discrepancy, which is important for augmented reality applications. Like MSSD, MSPD also measures the maximum distance instead of the average in order to be robust to the sampling density of vertices. MSPD Recall is the percent of MSPD scores less than $10$ thresholds ranging from $5$ to $50$ (measured in pixels).

\section{Additional Ablations}

\begin{table*}[t]
    \centering
    \resizebox{\textwidth}{!}{
    \begin{tabular}{lcccccc}
    \toprule
&\multicolumn{2}{c}{YCB-V \cite{ycb}}&\multicolumn{2}{c}{T-LESS \cite{tless}}&\multicolumn{2}{c}{LM-O\cite{linemod_occlusion}} \\
\cmidrule(lr){2-3}
\cmidrule(lr){4-5}
\cmidrule(lr){6-7}
&MSPD Recall&MSSD Recall&MSPD Recall&MSSD Recall&MSPD Recall&MSSD Recall \\
\midrule
\textbf{Bidirectional (depth as variable) PnP} & 0.833&\textbf{0.751}&\textbf{0.649}&\textbf{0.474}&\textbf{0.793}&\textbf{0.609} \\
Unidirectional (depth as variable) PnP [render to image]&\textbf{0.834}&0.750&0.639&0.456&0.792&0.593 \\
Unidirectional (depth as variable) PnP [image to render]&0.750&0.630& 0.480 & 0.234 & 0.645 & 0.379\\
\midrule
\textbf{Multiview renders} & \textbf{0.833}&\textbf{0.751}&\textbf{0.649}&\textbf{0.474}&\textbf{0.793}&\textbf{0.609} \\
Single render&0.814&0.727&0.619&0.429&0.772&0.569 \\
\midrule
\textbf{Predicting per-pixel confidence weights}&\textbf{0.833}&\textbf{0.751}&\textbf{0.649}&\textbf{0.474}&\textbf{0.793}&\textbf{0.609} \\
Uniform confidence&0.694&0.598&0.568&0.377&0.765&0.568 \\

\midrule
\textbf{Revised approach (depth as variable)} &\textbf{0.833}&\textbf{0.751}&\textbf{0.649}&\textbf{0.474}&\textbf{0.793}&\textbf{0.609} \\
Depth as variable but do not discard depth update & 0.744&0.674&0.520&0.326&0.725&0.497 \\
\midrule
\textbf{Use Gradient Clipping}&0.833&0.751&0.649&0.474&0.793&0.609 \\
No Gradient Clipping&\multicolumn{5}{c}{Training diverges causing NaNs}& \\
\midrule
\textbf{Bidirectional context features}&0.833&0.751&\textbf{0.649}&\textbf{0.474}&0.793&\textbf{0.609} \\
Using unidirectional context features&\textbf{0.848}&\textbf{0.764}&0.647&0.452&\textbf{0.807}&0.608 \\
\midrule
\textbf{Pose + Flow Loss}&\textbf{0.833}&\textbf{0.751}&\textbf{0.649}&\textbf{0.474}&\textbf{0.793}&\textbf{0.609} \\
Flow Loss Only &0.760&0.666&0.589&0.378&0.741&0.521 \\
Pose Loss Only &0.246&0.190&0.263&0.166&0.271&0.093 \\

\midrule
\textbf{Depth-augmented PnP (predicting depth revisions)} & \textbf{0.842}&\textbf{0.765}&0.647&0.465&\textbf{0.803}&\textbf{0.616} \\
No depth augmentation (no depth revisions) &0.833&0.751&\textbf{0.649}&\textbf{0.474}&0.793&0.609 \\
\bottomrule
    \end{tabular}}
    \caption{Additional ablation experiments using our method for RGB-Only input. We evaluate our method on a held-out subset of training images. Initial poses are generated by randomly perturbing the ground truth pose. Options used in our full RGB method are bolded.}
    \label{tab:ablations_rgb}
\end{table*}

\begin{table*}[t]
\centering
\resizebox{\linewidth}{!}{
\begin{tabular}{lcccccc}
\toprule
&\multicolumn{2}{c}{YCB-V \cite{ycb}}&\multicolumn{2}{c}{T-LESS \cite{tless}} &\multicolumn{2}{c}{LM-O \cite{linemod_occlusion}} \\
\cmidrule(lr){2-3}
\cmidrule(lr){4-5}
\cmidrule(lr){6-7}
&MSPD Recall&MSSD Recall&MSPD Recall&MSSD Recall&MSPD Recall&MSSD Recall \\
\midrule

\textbf{Bidirectional context features} &\textbf{0.924}&\textbf{0.955}&\textbf{0.685}&\textbf{0.582}&\textbf{0.828}&\textbf{0.788} \\
Unidirectional context features&0.910&0.944&\textbf{0.685}&\textbf{0.582}&{0.826}&{0.784} \\

\bottomrule
\end{tabular}
}
\caption{Additional ablation experiments using our method for RGB-D input. We evaluate our method on held-out training images. Initial poses are generated by randomly perturbing the ground truth pose.  Options used in our full method are bolded.}
\label{tab:ablations_rgbd}
\end{table*}

\noindent\textbf{Gradient Clipping} Treating depth as a variable in the optimization leads to unstable behavior after several thousand training steps. We solve this problem by clipping the gradients of the input to the GRU to a maximum of $0.01$ in the middle of the backward pass.

\smallskip
\noindent \textbf{Discarding Depth Update} In the RGB setting, we jointly optimize the pose and depth together. We then discard the depth update and generate a new depth map by rendering the updated pose (see Sec \ref{sec:rgb_optim}). This works better than jointly optimizing pose and depth together but applying the depth update to the previous estimate instead of discarding it.

\smallskip
\noindent\textbf{RGB-Only Bidirectional (depth as variable) PnP}
In the RGB setting, bidirectional PnP is helpful on T-LESS while benign on LM-O and YCB-V. On YCB-V, our method converges to an accurate pose even when only using correspondences from the input image to the rendered image (Tab. \ref{tab:ablations_rgb}). This indicates that the induced flow is sufficiently accurate even when using a depth \textit{approximation} in the mapping function $\Pi$ . 

\smallskip
\noindent\textbf{RGB-Only Depth-augmented PnP} In the RGB setting, depth residuals are helpful on YCB-V and LM-O but not on T-LESS (Tab. \ref{tab:ablations_rgb}). The inconsistent benefit could arise from the fact that the depth in the image is generated from the current pose estimate, and therefore predicting the appropriate depth residuals in the GRU is more challenging.

\smallskip
\noindent\textbf{Bidirectional Context Features} There is a significant domain gap between the input images and the rendered images. This is due to, among other things, innaccurate meshes resulting from imperfect scans, significant lighting differences and different reflectance properties. We evaluate the importance of extracting contextual features from both images, contrary to RAFT \cite{teed2020raft} which operates in a single image domain. Tabs. \ref{tab:ablations_rgb} \& \ref{tab:ablations_rgbd} show that using contextual information from only the source image is sufficient to establish accurate correspondences.

\begin{figure}[h]%
    \centering
    \includegraphics[width=1.0\textwidth]{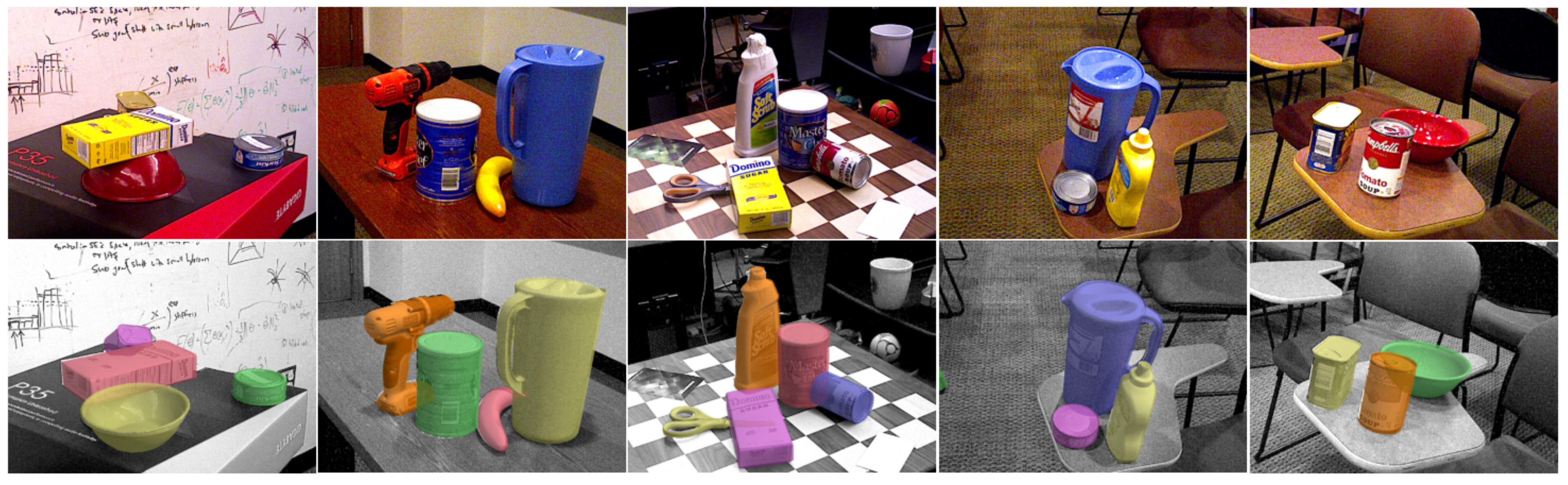} 
    \caption{Additional qualitative results on the YCB-V Dataset. Our method incorrectly orients the bowl and spam in the far left image.}%
    \label{fig:example}%
\end{figure}

\begin{figure*}[b]%
    \centering
    \subfloat[\centering Linemod (Occluded) Accuracy]{{\includegraphics[width=0.35\textwidth]{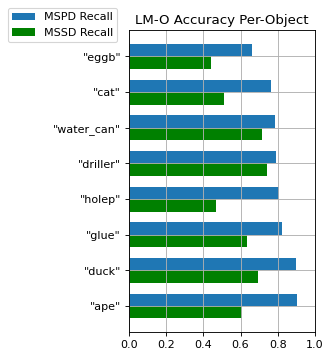} }}%
    \qquad
    \subfloat[\centering YCB-V Accuracy]{{\includegraphics[width=0.35\textwidth]{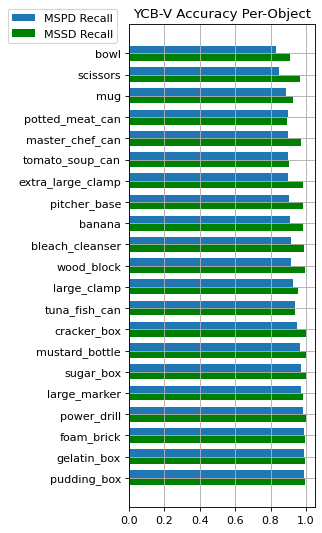} }}%
    \caption{Accuracy Per-Object on the Linemod (Occluded) and YCB-V Datasets}%
    \label{fig:example}%
\end{figure*}

\clearpage

\begin{figure}[t]%
    \centering
    \includegraphics[width=1.0\textwidth]{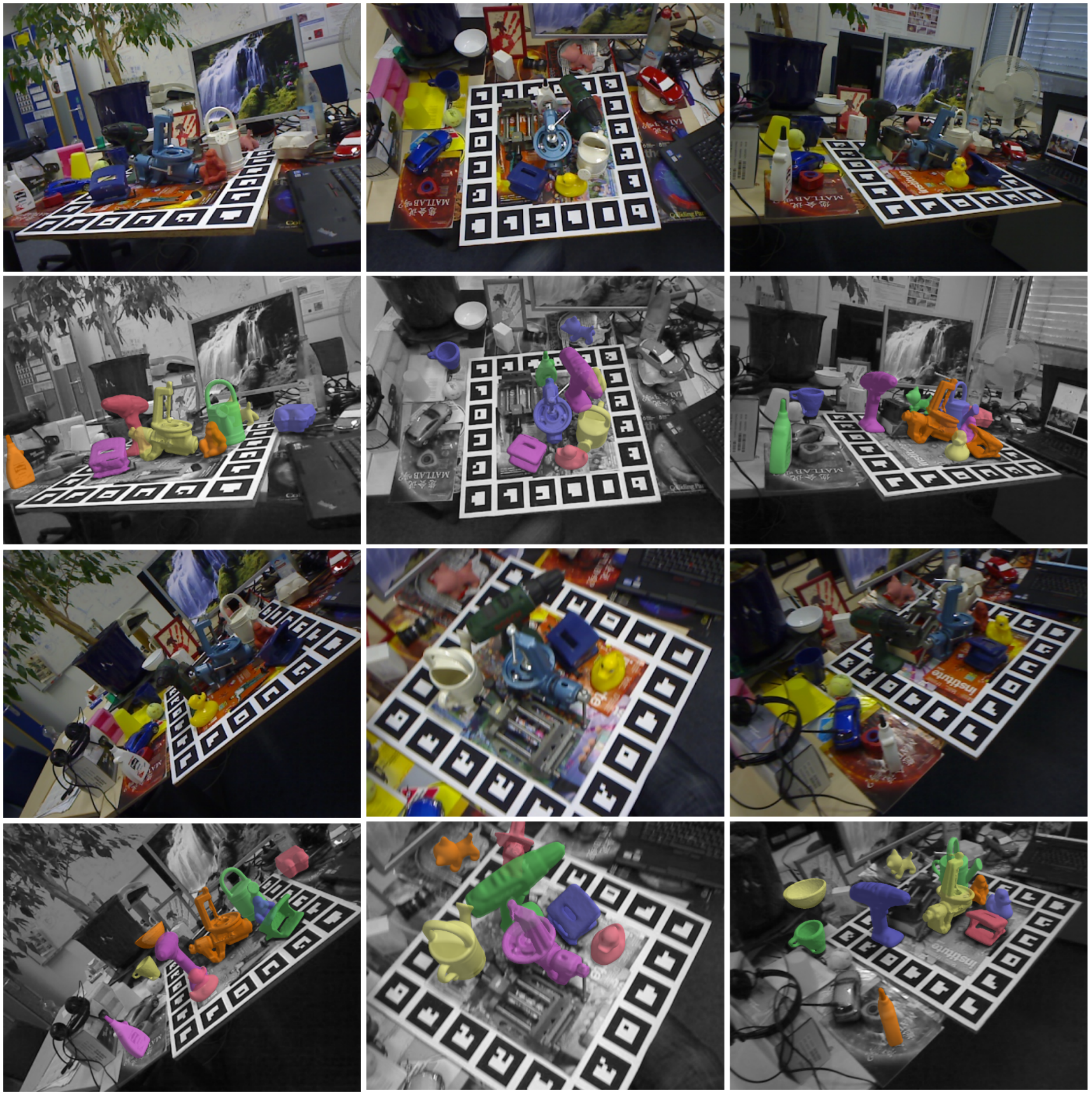} 
    \caption{Qualitative results on the Linemod (Occluded) Dataset}%
    \label{fig:example}%
\end{figure}

\clearpage

\begin{figure}[t]%
    \centering
    \includegraphics[width=1.0\textwidth]{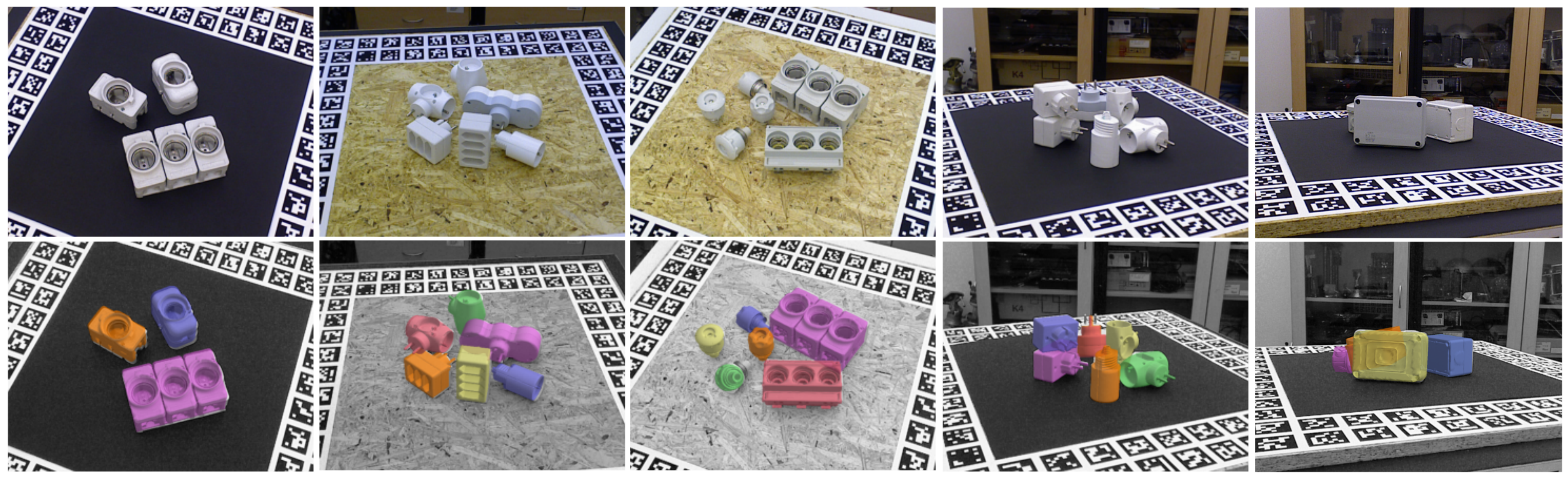} 
    \caption{Additional qualitative results on the T-LESS Dataset. Our method incorrectly positions the purple object in the far right image.}%
    \label{fig:example}%
\end{figure}

\noindent \textbf{Licenses:} Linemod and T-LESS licensed under CC BY 4.0. Linemod (Occluded) licensed under CC BY-SA 4.0. YCB-V licensed under MIT. Cosypose models licensed under MIT.

\end{document}